\documentclass[sigconf]{acmart}
\AtBeginDocument{%
  \providecommand\BibTeX{{%
    \normalfont B\kern-0.5em{\scshape i\kern-0.25em b}\kern-0.8em\TeX}}}

\setcopyright{acmcopyright}
\copyrightyear{2020}
\acmYear{2020}
\acmDOI{10.1145/3422844.3423051}
\acmConference[MMSports'20]{3rd International Workshop on Multimedia Content Analysis in Sports}{October 16, 2020}{Seattle, WA, USA}
\acmPrice{15.00}
\acmISBN{978-1-4503-8149-9/20/10}

\begin{document}

\title{SoccerDB: A Large-Scale Database for Comprehensive Video Understanding}


\author{
Yudong Jiang
}
\affiliation
{State Key Laboratory of Media Convergence Production Technology and Systems \& Xinhua Zhiyun Technology Co., Ltd.
}
\email
{nebuladream@gmail.com}

\author{
Kaixu Cui
}
\affiliation
{State Key Laboratory of Media Convergence Production Technology and Systems \& Xinhua Zhiyun Technology Co., Ltd.
}
\email
{cuikaixu@shuwen.com}

\author{
Leilei Chen
}
\affiliation
{State Key Laboratory of Media Convergence Production Technology and Systems \& Xinhua Zhiyun Technology Co., Ltd.
}
\email
{chenleilei@shuwen.com}

\author{
Canjin Wang
}
\affiliation
{State Key Laboratory of Media Convergence Production Technology and Systems \& Xinhua Zhiyun Technology Co., Ltd.
}
\email
{wangcanjin@shuwen.com}

\author{
Changliang Xu
}
\affiliation
{State Key Laboratory of Media Convergence Production Technology and Systems \& Xinhua Zhiyun Technology Co., Ltd.
}
\email
{xuchangliang@shuwen.com}

\renewcommand{\shortauthors}{Yudongjiang, et al.}
\begin{abstract}
Soccer videos can serve as a perfect research object for video understanding because soccer games are played under well-defined rules while complex and intriguing enough for researchers to study. In this paper, we propose a new soccer video database named SoccerDB, comprising 171,191 video segments from 346 high-quality soccer games. The database contains 702,096 bounding boxes, 37,709 essential event labels with time boundary and 17,115 highlight annotations for object detection, action recognition, temporal action localization, and highlight detection tasks. To our knowledge, it is the largest database for comprehensive sports video understanding on various aspects. We further survey a collection of strong baselines on SoccerDB, which have demonstrated state-of-the-art performances on independent tasks. Our evaluation suggests that we can benefit significantly when jointly considering the inner correlations among those tasks. We believe the release of SoccerDB will tremendously advance researches around comprehensive video understanding. {\itshape  Our dataset and code published on https://github.com/newsdata/SoccerDB.}
\end{abstract}




\maketitle

\section{Introduction}
Comprehensive video understanding is a challenging task in computer vision. It has been explored in ways including action recognition, temporal action localization, object detection, object tracking and so on. However, most works on video understanding mainly focus on isolated aspects of video analysis, yet ignore the inner correlation among those tasks. 

There are many obstacles for researchers doing the correlation study: first, the manual annotation of multiple tasks’ labels on a large-scale video database is extremely time-consuming; second, different approaches lack a fair and uniform benchmark excluding interference factors for conducting rigorous quantitative analysis; third, some datasets focusing on the areas that are not challenging or valuable enough to attract researchers’ attention.  We need research objects, which are challenging and with clear rules and restrictive conditions for us to conduct an accurate study on questions we are interested in. In this paper, we choose soccer matches as our research object, and construct a dataset with multiple visual understanding tasks featuring various analysis aspects, aiming at building algorithms that can comprehensively understand various aspects of videos like a human.

\begin{figure}
\centering
 \includegraphics[width=\linewidth]{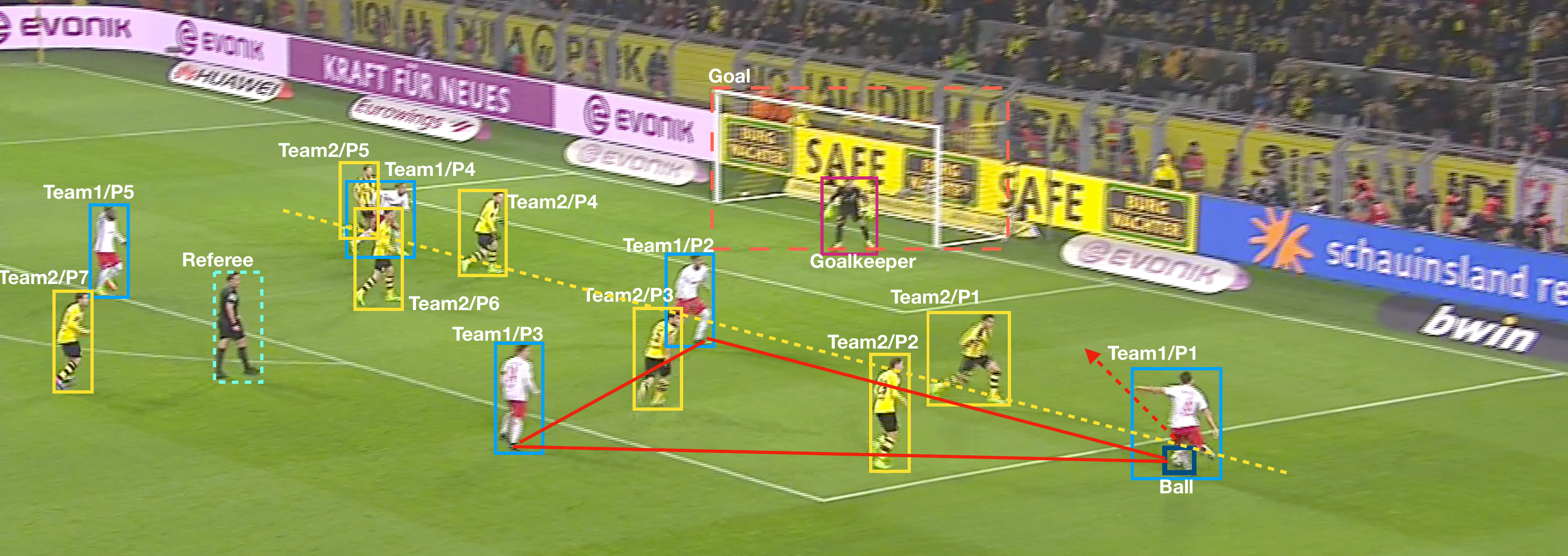}
 \caption{Soccer tactics visualization powered by object detection}
\label{fig:detection}
\end{figure}

\subsection{Soccer Video Understanding}
Soccer video understanding is not only valuable to academic communities but also lucrative in the commercial world. The European soccer market generates annual revenue of \$28.7 billion~\cite{giancola2018soccernet}. Regarding soccer content production, automatic soccer video analysis can help editors to produce match summaries, visualize key players' performance for tactical analysis, and so on. Some pioneering companies like GameFace\footnote{\url{http://gameface.ai/soccer}}, SportLogiq\footnote{\url{https://sportlogiq.com/en/technology}} adopt this technology on match statistics to analyze strategies and players’ performance. However, automatic video analysis has not fully met the markets' needs. The CEO of Wyscout, claims the company employs 400 people on soccer data, each of whom takes over 8 hours to provide up to 2000 annotations per game~\cite{giancola2018soccernet}.

\subsection{Object Detection}
Object detection has seen huge development over the past few years and gained human-level performance in applications including face detection, pedestrian detection, etc. To localize instances of semantic objects in images is a fundamental task in computer vision. 
In soccer video analysis, a detection system can help us to find positions of the ball, players, and goalposts on the field. With the position information, we can produce engaging visualization as shown in Figure \ref{fig:detection} for tactic analysis or enhance the fan experience. Though many advanced detection systems can output reliable results under various conditions, there are still many challenges when the object is small, fast-moving, or blur. In this work, we construct a soccer game object detection dataset and benchmark two state-of-the-art detection models under different framework: RetinaNet~\cite{lin2017focal}, a “one-stage” detection algorithm, and Faster R-CNN~\cite{ren2015faster}, a “two-stage” detection algorithm.

\subsection{Action Recognition}
Action recognition is also a core video understanding problem and has achieved a lot over the past few years. Large-scale datasets such as Kinetics~\cite{carreira2017quo}, Sports-1M~\cite{karpathy2014large}, YouTube-8M~\cite{abu2016youtube} have been published. Many state-of-the-art deep learning-based algorithms like I3D~\cite{carreira2017quo}, Non-local Neural Networks~\cite{wang2018non}, slowFast Network~\cite{feichtenhofer2019slowfast}, were proposed to this task. While supervised learning has shown its power on large scale recognition datasets, it failed when lacking training data. In soccer games, key events such as penalty kicks, are rare, which means many state-of-the-art recognition models cannot output convincing results when facing these tasks. We hope this problem could be further investigated by considering multiple objects' relationships as a whole in the dataset.

\begin{figure}
\centering
   \includegraphics[width=\linewidth]{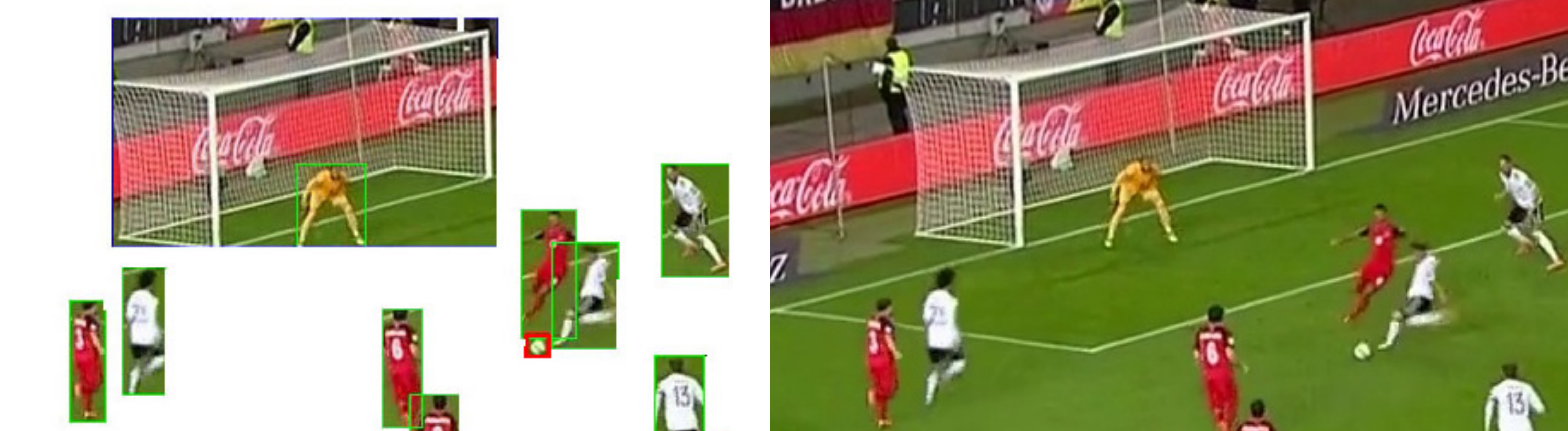}
   \caption{The moment of a shooting. Right side: original image. Left side: only keep players, ball, goal areas in image}
\label{fig:shot}
\end{figure}

In this paper, we also provide our insight into the relationship between object detection and action recognition. We observe that since soccer match supply simplex scene and object classes, it is extraordinarily crucial to model the special relationships of objects and their change over time. Imagine, if you can only see players, the ball and goal posts from a game's screenshot, could you still understand what is happening on the field? Look at the left picture in Figure \ref{fig:shot}, maybe you have guessed right: {\em that's the moment of a shooting}. Although modeling human-object or object-object interactions have been explored to improve action recognition~\cite{gkioxari2015contextual}~\cite{wang2018videos} in recent years, we still need to have a closer look at how do we use the detection knowledge boosting action recognition more efficiently? Our experiments show that the performance of a state-of-the-art action recognition algorithm can be increased by a large margin while combining with object class and location knowledge.

\subsection{Temporal Action Localization}
Temporal action localization is a significant and more complicated problem than action recognition in video understanding because it requires to recognize both action categories and the time boundary of an event. The definition of the temporal boundary of an event is ambiguous and subjective, for instance, some famous databases like Charades and MultiTHUMOS are not consistent among different human annotators~\cite{sigurdsson2017actions}. This also increases our difficulty when labeling for SoccerDB. To overcome the challenge of ambiguity, we define soccer events with a particular emphasis on time boundaries, based on the events' actual meaning in soccer rules. For example, we define {\em red/yellow card} as starting from a referee showing the card, and ending when the game resuming. The new definition helps us to get more consist of action localization annotations.

\subsection{Highlight Detection}
The purpose of highlight detection is to distill interesting content from a long video. Because of the {\em subjectivity problem}, to construct a highlight detection dataset usually requires multi-person labeling for the same video. It will greatly increase the costs and limit the scale of the dataset~\cite{song2015tvsum}. We find in soccer TV broadcasts, video segments containing highlight events are usually replayed many times, which can be taken as an important clue for soccer video highlight detection. Many works explored highlight detection while considering replays. Zhao Zhao et. al proposed a highlight summarization system by modeling Event-Replay(ER) structure~\cite{zhao2006highlight}, A. Raventós et. al used audio-visual descriptors for automatic summarization which introduced replays for improving the robustness~\cite{raventos2015automatic}. SoccerDB provides a {\em playback} label and reviews this problem by considering the relationship between the actions and highlight events.

\subsection{Contributions}
\begin{itemize}
\item We introduce a challenging database on comprehensive soccer video understanding. Object detection, action recognition, temporal action localization, and highlight detection. Those tasks, crucial to video analysis, can be investigated in the closed-form under a constrained environment.
\item We provide strong baseline systems on each task, which are not only meant for academic researches but also valuable for automatic soccer video analysis in the industry.
\item We discuss the benefit when considering the inner connections among different tasks: we demonstrate modeling objects' spatial-temporal relationships by detection results could provide complementary representation to the convolution-based model learned from RGB that increases the action recognition performance by a large margin; joint training on action recognition and highlight detection can boost the performance on both tasks.
\end{itemize}

\section{Related Work}
\subsection{Sports Analytics}
Automated sports analytics, particularly those on soccer and basketball, are popular around the world. The topic has been profoundly researched by the computer vision community over the past few years. Vignesh Ramanathan et al. brought a new automatic attention mechanism on RNN to identify who is the key player of an event in basketball games~\cite{ramanathan2016detecting}. Silvio Giancola et al. focused on temporal soccer events detection for finding highlight moments in soccer TV broadcast videos~\cite{giancola2018soccernet}. Rajkumar Theagarajan et al. presented an approach that generates visual analytics and player statistics for solving the talent identification problem in soccer match videos~\cite{theagarajan2018soccer}. Huang-Chia Shih surveyed 251 sports video analysis works from content-based viewpoint for advancing broadcast sports video understanding~\cite{shih2017survey}. The above works were only the tip of the iceberg among magnanimous research achievements in the sports analytics area. 

\subsection{Datasets}\label{sec:datasets}
Many datasets have been contributed to sports video understanding. Vignesh Ramanathan et al. provided 257 basketball games with 14K event annotations corresponding to 10 event classes for event classification and detection~\cite{ramanathan2016detecting}. Karpathy et al. collected one million sports videos from Youtube belonging to 487 classes of sports promoting deep learning research on action recognition greatly~\cite{karpathy2014large}. 
Datasets for video classification in the wild have played a vital role in related researches. Two famous large-scale datasets, Youtube-8M~\cite{abu2016youtube} and Kinetics~\cite{carreira2017quo} were widely investigated, which have inspired most of the state-of-the-art methods in the last few years. Google proposed the AVA dataset to tackle the dense activity understanding problem, which contained 57,600 clips of 3 seconds duration taken from featured films~\cite{gu2018ava}. ActivityNet explored general activity understanding by providing 849 video hours of 203 activity classes with an average of 137 untrimmed videos per class and 1.41 activity instances per video\cite{caba2015activitynet}. Although ActivityNet considered video understanding from multiple aspects including semantic ontology, trimmed and untrimmed video classification, spatial-temporal action localization, we argued that it is still too far away from a human-comparable general activity understanding in an unconstrained environment. Part of the source videos in our dataset was collected from SoccerNet~\cite{giancola2018soccernet}, a benchmark with a total of 6,637 temporal annotations on 500 complete soccer games from six main European leagues. A comparison between different databases is shown in Table \ref{tab:dbcompare}.

\section{Creating SoccerDB}
\subsection{Object Detection Dataset Collection}
To train a robust detector for different scene, we increase the diversity of the dataset by collecting data from both images and videos. We crawl 24,475  images of soccer matches from the Internet covering as many different scenes as possible, then use them to train a detector for boosting the labeling process. For video parts, we collect 103 hours of soccer match videos including 53 full-match and 18 half-match which source is described in section \ref{sec:vidcol}. To increase the difficulty of the dataset, we auto-label each frame from the videos by the detector trained on image set, then select the keyframes with poor predictions as the dataset proposals. Finally, we select 45,732 frames from the videos for object detection task. As shown in Table \ref{tab:detbox}, the total number of bounding box labels for image parts are 142,579, with 117,277 player boxes, 19,072 ball boxes, and 6,230 goal boxes, the total number of bounding box labels for video parts are 702,096, with 643,581 player boxes, 45,160 ball boxes, and 13,355 goal boxes. We also calculate the scale of the boxes by COCO definition~\cite{lin2014microsoft}. The image parts is spited into 21,985 images for training, and 2,490 for testing randomly. For the video parts random select 18 half-matches for testing, the other matches for training yielding 38,784 frames for training and 6,948 for testing. 
\begin{table}
\caption{Bounding box statistics for object detection dataset. The scale of the bounding box, small, medium and large, following the definition of COCO dataset. *-img represent the image parts, *-vid represent the video parts}
\label{tab:detbox}
\begin{tabular}{|l|r|r|r|r|}
\hline                  
Classes & \#Small & \#Medium  & \#Large & \#All \\
\hline                  
Player-img  & 27409 & 50231 & 39637 & 117277  \\
Ball-img  & 7261  & 7131  & 4680  & 19072 \\
Goal-img  & 51  & 922 & 5257  & 6230  \\
\hline                  
Total-img & 34721 & 58284 & 49574 & 142579  \\
\hline                  
Player-vid  & 164716  & 468335  & 10530 & 643581  \\
Ball-vid  & 43027 & 2066  & 67  & 45160 \\
Goal-vid  & 138 & 2631  & 10586 & 13355 \\
\hline                  
Total-vid & 207881  & 473032  & 21183 & 702096  \\
\hline                  
Total & 242602  & 531316  & 70757 & 844675  \\
\hline                  
\end{tabular}
\end{table}

\subsection{Video Dataset Collection}\label{sec:vidcol}
We adopt 346 high-quality full soccer matches' videos, including 270 matches from SoccerNet~\cite{giancola2018soccernet} covering six main European leagues ranging from 2014 to 2017 three seasons, 76 matches videos from the China Football Association Super League from 2017 to 2018, and the 18th, 19th, 20th FIFA World Cup\footnote{We will provide the mapping table of the annotated videos in SoccerDB and SoccerNet. Researchers need to apply the SoccerNet to get the video. The other videos need to comply with the non-disclosure agreement, similar to the SoccerNet protocol.}. The whole dataset consumes 1.4 TB storage, with a total duration of 668.6 hours. We split the games into 226 for training, 63 for validation, and 57 for testing randomly. {\em All videos for object detection are not included in this video dataset.}

\subsection{Event Annotations}
We define ten different soccer events which are usually the highlights of the soccer game with standard rules for their definition. We define the event boundaries as clear as possible and annotate all of them densely in long soccer videos. The annotation system records the start/end time of an event, the categories of the event and if the event is a playback. An annotator takes about three hours to label one match, and another experienced annotator reviews those annotations to ensure the outcomes' quality.

\subsection{Video Segmentation Processing}
We split the dataset into 3 to 30 seconds segments for easier processing. We make sure an event would not be divided into two segments, and keep the event's temporal boundary in one segment. Video without any event is randomly split into 145,473 video clips with time duration from 3 to 20 seconds. All of the processed segments are checked again by humans to avoid annotation mistakes. Some confusing segments are discarded during this process. Finally, we get a total of 25,719 video segments with event annotations (core dataset) and 145,473 background segments. There are 1.47 labels per segment in the core dataset.

\newcommand{\tabincell}[2]{\begin{tabular}{@{}#1@{}}#2\end{tabular}}  
\begin{table}
\begin{center}
\caption{SoccerDB statistics. The dataset covers ten key events in soccer games. This table shows segment number, total time duration and playback segment number of each events. The unit of the duration is minute}
\label{tab:soccerdb}
\begin{tabular}{|l|r|r|r|r|}
\hline
Events & \#Segments & Dur(min) &  \#Playback  \\
\hline
Background(BK) & 145473 & 25499.3 & 0 \\
Injured(IJ) & 1478 & 306.57 & 666 \\
Red/Yellow Card(R/Y) & 1160 & 244.08 & 219 \\
Shot(SH) & 14358 & 2125.35 & 8490 \\
Substitution(SU) & 867 & 298.92 & 14 \\
Free Kick(FK) & 3119 & 400.53 & 843  \\
Corner(CO) & 3275 & 424.08 & 668  \\
Saves(SA) & 5467 & 735.95 & 2517 \\
Penalty Kick(PK) & 156 & 28.25 & 130  \\
Foul(FO) & 5276 & 766.33 & 1015   \\
Goal(GO) & 2559 & 532.03 & 2559 \\
\hline
Total & 183188 & 31361.39 & 17121 \\
\hline
\end{tabular}
\end{center}
\end{table}

\subsection{Dataset Analysis}
Details of SoccerDB statistics are shown in Table \ref{tab:soccerdb}. A total of 14,358 segments have {\em shot} labels, which account for 38.07\% among all events, except for the background. In contrast, we only collected 156 segments for {\em penalty kick}, and 1160 for {\em red and yellow card}, accounting for 0.41\% and 3.07\%, respectively. Since the dataset has an extreme class imbalance problem, it is difficult for the existing state-of-the-art supervised methods to produce convincing results. We also explored the distribution of playbacks and found its relevance to events' type, as every goal event has playbacks, contrasting with only 1.6\% proportion of {\em substitution} have playbacks. In section \ref{sec:highlight} we prove this relevance. As shown in section \ref{sec:datasets}, we also provide comparisons of many aspects between other popular datasets and ours. Our dataset supports more variety in tasks and more detailed soccer class labels for constrained video understanding.

\begin{table*}
\begin{center}
\caption{The comparison of different datasets on video understanding. In support tasks column [1]: Video Classification, [2]: Spatial-Temporal Detection, [3]: Temporal Detection, [4]: Highlight Detection [5]: Object Detection. The background is taken as a class in classes number statistics  }
\label{tab:dbcompare}
\begin{tabular}{|l|l|r|r|r|r|c|}
\hline
Datasets    &  Context    &     \#Video    &     \#Instance    &     Dur(hrs) &     \#Classes    &     Support Tasks    \\
\hline
YouTube-8M    &    General    &    6100000    &    18300000    &    350000    &    3862    &    [1]    \\
\hline
Kinetics-600    &    General    &    495547    &    495547    &    1377    &    600    &    [1]    \\
\hline
AVA dataset    &    Movies    &    57600    &    210000    &    48    &    80    & [1][2]\\
\hline
ActivityNet    &    General    &    19994    &    30791    &    648    &    200    & [1][3] \\
\hline
Sports-1M    &    Sports    &    1133158    &    -    &    -    &    487    &    [1]    \\
\hline
SoccerNet    &    Soccer    &    1000    &    6637    &    764    &    4    & [1][3] \\
\hline
NCAA    & Basketball    &    257    &    14000    &    385.5    &    11    & [1][3] \\
\hline
SoccerDB    &    Soccer    &    171191    &    37715    &    668.6    &    11    & [1][3][4][5] \\
\hline
\end{tabular}
\end{center}
\end{table*}

\section{The Baseline System}

To evaluate the capability of current video understanding technologies, and also to understand challenges to the dataset, we developed algorithms that have show strong performances on various datasets, which can provide strong baselines for future work to compare with. {\em In our baseline system, the action recognition sub-module plays an essential role by providing basic visual representation to both temporal action detection and highlight detection tasks.}

\subsection{Object Detection}\label{sec:obj}
We adopt two representative object detection algorithms as baselines. One is Faster R-CNN, developed by Shaoqing Ren et al.~\cite{ren2015faster}. The algorithm and its variant are widely used in many detection systems in recent years. Faster R-CNN belongs to the two-stage detector: The model using RPN proposes a set of regions of interests (RoI), then a classifier and a regressor only process the region candidates to get the category of the RoI and precise coordinates of bounding boxes. Another one is RetinaNet, which is well known as a one-stage detector. The authors Tsung-Yi Lin et al. discover that the extreme foreground-background class imbalance encountered is the central cause and introduced focal loss for solving this problem~\cite{lin2017focal}.

\subsection{Action Recognition}\label{sec:actionReco}
We treat each class as a binary classification problem. Cross entropy loss is adopted for each class. Two state-of-the-art action recognition algorithms are explored, the slowFast Networks and the Non-local Neural Networks. The slowFast networks contain two pathways: a slow pathway, simple with low frame rate, to capture spatial semantics, and a fast pathway, opposed to the slow pathway, operating at a high frame rate, to capture the motion pattern. We use ResNet-50 as the backbone of the network. The Non-local Neural Networks proposed by Xiaolong Wang et. al~\cite{wang2018non}, that can capture long-range dependencies on the video sequence. The non-local operator as a generic building block can be plugged into many deep architectures. We adopt I3D with ResNet-50 backbone and insert non-local operators.

\begin{figure}[t]
\begin{center}
  \includegraphics[width=1.0\linewidth]{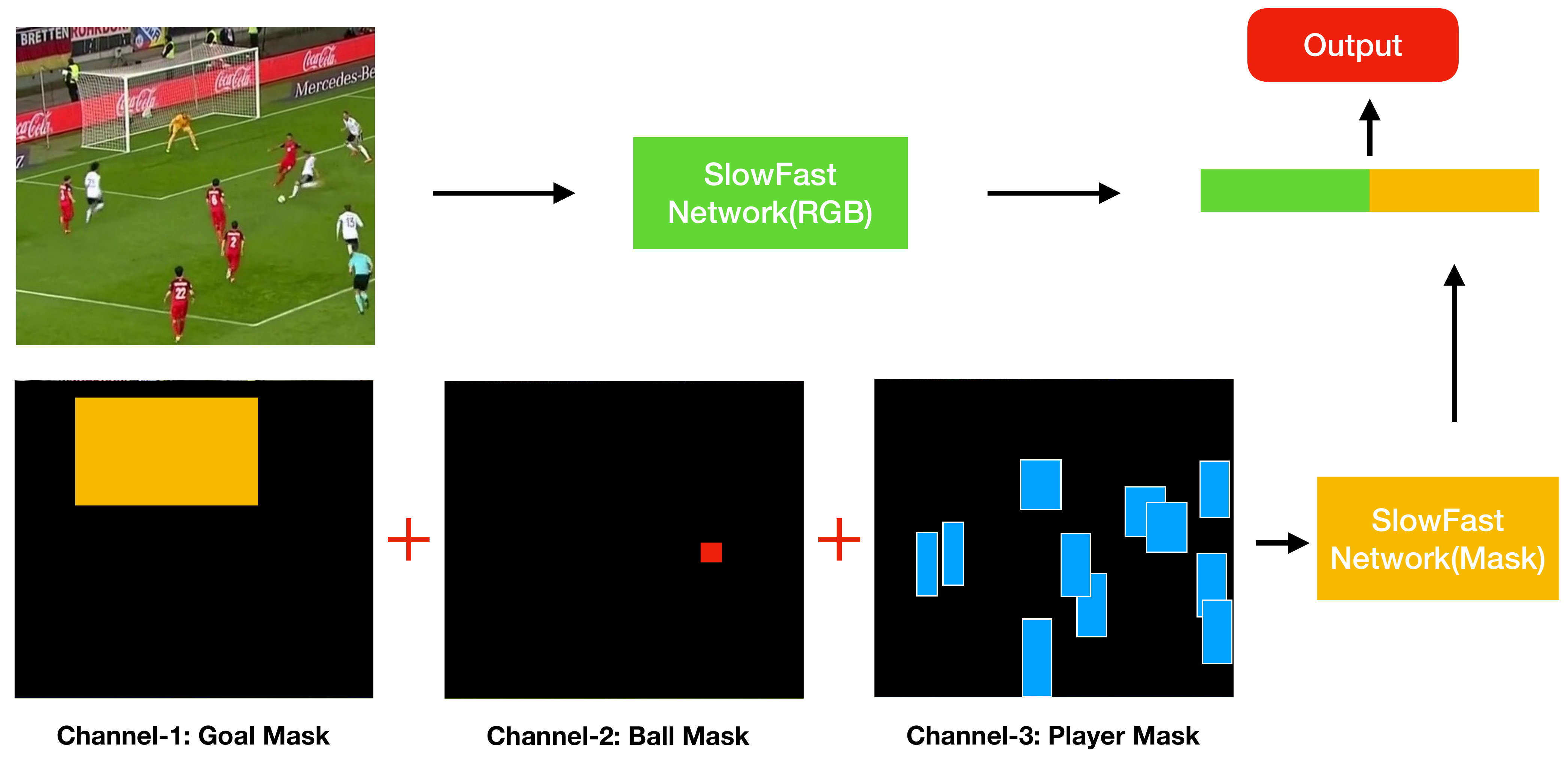}
\end{center}
  \caption{Mask and RGB Two-Stream (MRTS) approach structure.}
\label{fig:transferlearning}
\end{figure}
\subsection{Transfer Knowledge from Object Detection to Action Recognition}\label{sec:actionRecoObj}
We survey the relationship between object detection and action recognition based on Faster R-CNN and SF-32 network (slowFast framework by sampling 32 frames per video segments) mentioned in section \ref{sec:obj} and \ref{sec:actionReco}. First, we use Faster R-CNN to detect the objects from each sampled frame. Then, as shown in figure \ref{fig:transferlearning}, we add a new branch to SF-32 for modeling object spatial-temporal interaction explicitly for explaning: object detection can provide complementary objects interaction knowledge that convolution-based model could not learn from the RGB sequence.

{\em Mask and RGB Two-Stream (MRTS) approach.} We generate object masks as the same size of the RGB frame, the channel size of the mask is equal to the object class number. For each channel, representing that one object class, the areas containing objects belong to this class are set to 1, others are set to 0. we set a two-stream ConvNet architecture, one stream takes the mask as input, the other input original RGB frame. Two streams are converged by concatenating the last fully-connected layers. We suppose that if the spatial-temporal modeling of object location can provide complementary representation, the result of this approach exceeds the baseline SF-32 network performance largely. 

\subsection{Temporal Action Detection}
We explore temporal action detection by a two-stage based method. First, a class-agnostic algorithm generates potential event proposals, then apply the classifying proposals approach for final temporal boundary localization. During the first stage, we utilize Boundary-Matching Network (BMN), a “bottom-up” temporal action proposal generation method, for generating high-quality proposals~\cite{lin2019bmn}. The BMN model is composed of three modules: (1) {\em Base module} processes the extracted feature sequence of the origin video, and output another video embedding shared by {\em Temporal Evaluation Module (TEM)} and {\em Proposal Evaluation Module (PEM)}. (2) TEM evaluates the starting and ending probabilities of each location in a video to generate boundary probability sequences. (3) PEM transfers the feature to a boundary-matching feature map which contains confidence scores of proposals. During the second stage, an action recognition models mentioned in section \ref{sec:actionReco} predicts the classification score of each top K proposals. We choose the highest prediction score of each class as the final detection result.

\subsection{Highlight Detection}
In this section, we formalize the highlight detection task as a binary classification problem, to recognize which video is a {\em playback} video. The structures of the highlight detection models are presented in Figure \ref{fig:highlight}. We select SF-32 network as the basic classifier, then we consider four scenarios:

\begin{figure}[t]
\begin{center}
  \includegraphics[width=\linewidth]{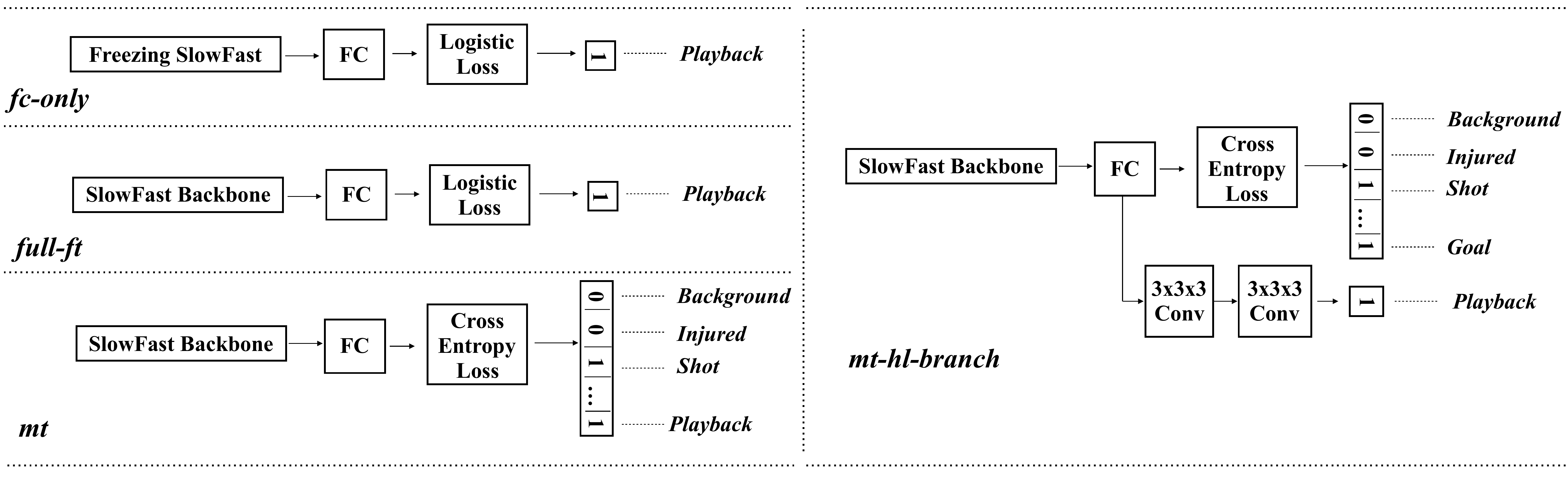}
\end{center}
  \caption{The structure of the highlight detection models}
\label{fig:highlight}
\end{figure}

\begin{itemize}
\item {\em Fully-connected only (fc-only) approach} involves extracting features from the final fc layer of a pre-trained model which is trained by action recognition task as shown in section \ref{sec:actionReco}. Then we train a logistic regressor for highlight detection. This approach evaluates the strength of the representation learned by action recognition, which can indicate the internal correlation between highlight detection and action recognition tasks.
\item {\em Fully Fine-tuning (full-ft) approach} fine-tuning a binary classification network by initializing weights from the action recognition model.
\item {\em Multi-task (mt) approach} we train a multi-label classification network for both action recognition and highlight detection tasks. We adopt a per-label sigmoid output followed by a logistic loss at the end of slowFast-32 network. This approach takes highlight segments as another action label in the action recognition framework. The advantage of this setting is that it can force the network to learn the relevance among tasks, while the disadvantage is that the new label may introduce noise confusing the learning procedure.
\item {\em Multi-task with highlight detection branch (mt-hl-branch) approach} we add a new two layers 3x3x3 convolution branch for playback recognition, which shares the same backbone with the recognition task. We only train the highlight detection branch by freezing action recognition pre-trained model initialized parameters at first, then fine-tune all parameters for multi-task learning.
\end{itemize}

\section{Experiments}
In this section, we focus on the performance of our baseline system on SoccerDB for object detection, action recognition, temporal action detection, and highlight detection tasks.

\subsection{Object Detection}
\begin{table}
\caption{The $AP_{0.5:0.95}$(\%) of RetinaNet and Faster R-CNN on different object scale. *-img represent the image parts, *-vid represent the video parts}
\label{tab:objDetRes}
\begin{tabular}{|l|r|r|r|r|r|}
\hline
Methods   &  small & medium & large & all \\
\hline
RetinaNet-img    &    30.2     &    63.1     &    75.4     &    64.8     \\
F.R-CNN-img    &    31.3     &    62.7     &    73.9     &    63.0     \\
\hline
RetinaNet-vid    &    39.9     &    57.3     &    61.8     &    62.3     \\
F.R-CNN-vid    &    42.5     &    58.1     &    58.8     &    62.0    \\
\hline
\end{tabular}
\end{table}

\begin{table}
\caption{The $AP_{0.5:0.95}$(\%) of RetinaNet and Faster R-CNN on different object classes. *-img represent the image parts, *-vid represent the video parts}
\label{tab:objDetClsRes}
 \begin{tabular}{|l|r|r|r|r|r|}
\hline
Methods &  mAP & player & ball & goal \\
\hline
RetinaNet-img    &    64.7     &    59.9     &    61.4     &    72.9     \\
F.R-CNN-img    &    63.0     &    57.5     &    59.9     &    71.7     \\
\hline
RetinaNet-vid    &    62.2     &    73.9     &    41.6     &    71.2     \\
F.R-CNN-vid    &    61.9     &    74.3     &    41.0     &    70.5     \\
\hline
\end{tabular}
\end{table}

We choose ResNeXt-101 with FPN as the backbone of both RetinaNet and Faster R-CNN. We use a pre-trained model on the MS-COCO dataset, and train the models by 8 NVIDIA-2080TI GPUs, with the initial learning rate of 0.01 for RetinaNet, and 0.02 for Faster R-CNN. MS-COCO style\footnote{\url{http://cocodataset.org/\#detection-eval}} evaluation method is applied to models' benchmark. The training data from both video parts and image parts are mixed to train each model. We present AP with IoU=0.5:0.95 and multi-scale in table \ref{tab:objDetRes}, and also report the AP of each class as shown in table \ref{tab:objDetClsRes}.
RetinaNet performs better than Faster R-CNN, and large-scale object is easier for both methods than the small object. The ball detection result is lower than the player and goal dual to the small scale and motion blur issue. All of the detection experiments are powered by mmdetection software which is developed by the winner of the COCO detection challenge in 2018~\cite{chen2019mmdetection}.

\begin{table*}
\begin{center}
\caption{Average precision(\%) of different recognition models on each classes. SF-32/SF-64: slowFast Network with 32/64 sample rates. NL-32/NL-64: Non-local Network with 32/64 sample rates. MRTS is the method powered by object detection. The events name for shot are the same as Table \ref{tab:soccerdb} }
\label{tab:actionReco}
\begin{tabular}{|l|r|r|r|r|r|r|r|r|r|r|r|r|}
\hline
Method    &    BK    &    IJ    &    R/Y    &    SH    &    SU    &    FK    &    CO    &    SA    &    PK    &    FO    &    GO    &    mAP    \\
\hline
SF-32    &    99.08     &    23.03     &    28.62     &    82.98     &    92.34     &    73.33     &    91.76     &    38.91     &    63.02     &    64.75     &    31.89     &    62.70     \\
NL-32    &    99.16     &    36.06     &    36.74     &    85.32     &    90.60     &    72.92     &    91.82     &    40.77     &    48.51     &    65.75     &    31.92     &    63.60     \\
SF-64    &    99.32     &    22.56     &    46.62     &    88.25     &    93.44     &    77.34     &    93.16     &    52.24     &    73.48     &    67.78     &    47.44     &    69.24     \\
NL-64    &    99.26     &    37.70     &    48.83     &    85.17     &    90.30     &    74.30     &    91.92     &    42.17     &    53.36     &    68.01     &    39.94     &    66.45     \\
MRTS    &    99.44     &    39.14     &    60.64     &    90.19     &    92.24     &    73.46     &    92.62     &    52.19     &    67.00     &    70.09     &    56.23     &    72.11     \\
\hline
\end{tabular}
\end{center}
\end{table*}
\subsection{Action Recognition}\label{sec:actionRecoExp}
We set up the experiments by open-source tool PySlowFast\footnote{\url{https://github.com/facebookresearch/SlowFast}}, and boost all recognition network from Kinetics pre-training model. Since some labels are rare in the dataset, we adjust the distribution of different labels appearing in the training batch to balance the proportion of labels. We resize the original video frames to 224x224 pixels and do horizontal flip randomly on the training stage. On the inference stage, we just resize the frame to 224x224 without a horizontal flip. We compare 32 and 64 sample frame number for investigating the sample rate influence. For each class, the average precision (AP) scores are demonstrated on Table \ref{tab:actionReco}.

The dense frame sample rate surpasses the sparse sample rate for both methods. The classes with more instances like {\em shot} perform better than classes with fewer instances. {\em Substitution} and {\em corner} with discriminative visual features to others, obtain high AP scores too. The AP of {\em penalty kick} fluctuates in a wide range because there are only 30 instances in the validation dataset.


\subsection{Transfer Knowledge from Object Detection to Action Recognition}
To make the results more comparable, all the basic experiment settings in this section are the same as described in section \ref{sec:actionRecoExp}.  The average precision results of MRST approach introduced by section \ref{sec:actionRecoObj} are shown in Table \ref{tab:actionReco}. 

From the experiment results, we can easily conclude that understanding the basic objects spatial-temporal interaction is critical for action recognition. MRST increases SF-32 by 15\%, which demonstrates the objects' relationship modeling can provide complementary representation that cannot be captured by 3D ConvNet from RGB sequence.

\subsection{Temporal Action Detection} 
In this section, we evaluate performances of temporal action proposal generation and detection and give quantified analysis on how action recognition task affects temporal action localization. For a fair comparison of different action detection algorithms, we benchmark our baseline system on the core dataset instead of the results produced by section \ref{sec:actionReco} models. We adopt the fc-layer of action classifier as a feature extractor on contiguous 32 frames getting 2304 length features. We set 32 frames sliding window with 5 frames for each step, which produces overlap segments for a video. The feature sequence is re-scaled to a fixed length D by zero-padding or average pooling with D=100. To evaluate proposal quality, Average Recall (AR) under multiple IoU thresholds [0.5:0.05:0.95] is calculated. We report AR under different Average Number of proposals (AN) as AR@AN, and the area under the AR-AN curve (AUC) as ActivityNet-1.3 metrics, where AN ranges from 0 to 100. To show the different feature extractor influence on the detection task, we compare two slowFast-32 pre-trained models, one is trained on the SoccerDB action recognition task described in section \ref{sec:actionReco}, another is trained on Kinetics. Table \ref{tab:actionDetection} demonstrates the results of those two extractors.

\begin{table}
\begin{center}
\caption{Temporal action proposal AR@AN(\%) and AUC(\%) results}
\label{tab:actionDetection}
\begin{tabular}{|l|r|r|r|r|r|r|}
\hline
Extractor    &    @1 & @10 & @50 & @100 & AUC \\
\hline
Kinetics    &    58.36    &    83.35    &    87.01    &    88.21    &    85.21    \\
SoccerDB    &    61.22    &    84.01    &    87.70    &    88.82    &    85.91    \\
\hline
\end{tabular}
\end{center}
\end{table}

The feature extractor trained on SoccerDB exceeds Kinetics extractor by 0.7\% on the AUC metric. {\em The results mean we benefit from training feature encoder on the same dataset on temporal action proposal generation stage, but the gain is limited.}  We use the same SF-32 classifier to produce the final detection results based on temporal proposals, and the detection metric is mAP with IoU thresholds \{0.3:0.1:0.7\}. For Kinetics proposals the mAP is 52.35\%, while SoccerDB proposals mAP is 54.30\%. The similar performance adopts by different feature encoder due to following reasons: first, Kinetics is a very large-scale action recognition database which contains ample patterns for training a good general feature encoder; second, the algorithm we adopt on proposal stage is strong enough for modeling the important event temporal location.

\begin{table}
\begin{center}
\caption{Highlight detection AP(\%) under four different setting}
\label{tab:highlightresult}
\begin{tabular}{|l|r|r|r|r|}
\hline
Methods    &  fc-only    &  full-ft &     mt    &  mt-hl-branch \\
\hline
AP    & 68.72    & 76.99 &    74.65 &    78.50 \\
\hline
\end{tabular}
\end{center}
\end{table}

\begin{table}
\begin{center}
\caption{Highlight detection multi-task learning mAP(\%) on action recognition. SF-32 is the baseline model }
\label{tab:highReco}
\begin{tabular}{|l|r|r|r|}
\hline
Methods    &    SF-32    &    mt    &    mt-hl-branch \\
\hline
mAP    &    62.70    &    60.86    &    64.16 \\
\hline
\end{tabular}
\end{center}
\end{table}
\subsection{Highlight Detection}\label{sec:highlight}
We set the experiments on the whole SoccerDB dataset. The average precision results of our four baseline models are shown in Table \ref{tab:highlightresult}. The fc-only model gets 68.72\% AP demonstrates the action recognition model can provide strong representation to highlight detection tasks indicating a close relationship between our defined events and the highlight segments. The mt model decreases the AP of the full-ft model by 2.33\%, which means the highlight segments are very different from action recognition when sharing the same features. The mt-hl-branch model gives the highest AP by better utilizing the correlation between the two tasks while distinguishing their differences. We also find the mt model is harmful to the recognition which decreases the mAP by 1.85 comparing to the baseline model. The mt-hl-branch can increase the action recognition mAP by 1.46\% while providing the highest highlight detection score. The detailed action recognition mAP for the three models is shown in Table \ref{tab:highReco}. A better way to utilize the connection between action recognition and highlight detection is expected to be able to boost the performances on both of them. 

\section{Conclusion}
In this paper, we introduce SoccerDB, a new benchmark for comprehensive video understanding. It helps us discuss object detection, action recognition, temporal action detection, and video highlight detection in a closed-form under a restricted but challenging environment. We explore many state-of-the-art methods on different tasks and discuss the relationship among those tasks. The quantified results show that there are very close connections between different visual understanding tasks, and algorithms can benefit a lot when considering the connections. We release the benchmark to the video understanding community in the hope of driving researchers towards building a human-comparable video understanding system.
\section{Acknowledgments}
This work is supported by State Key Laboratory of Media Convergence Production Technology and Systems, and Xinhua Zhiyun Technology Co., Ltd..

\bibliographystyle{ACM-Reference-Format}
\bibliography{main}

\end{document}